\documentclass[runningheads]{llncs}
\usepackage{cite}
\usepackage{xcolor,colortbl}
\usepackage{graphicx}
\usepackage{epstopdf}
\graphicspath{{eps/}}
\usepackage{upgreek}
\DeclareGraphicsExtensions{.eps}
\usepackage[cmex10]{amsmath}
\usepackage{amsfonts}
\usepackage{amssymb}
\usepackage{enumerate}
\usepackage{url}
\usepackage{multirow}
\usepackage{siunitx}
\usepackage{scrextend}
\usepackage{tabularx}
\usepackage{fixltx2e}
\usepackage{multirow}
\usepackage{subfigure}
\usepackage{floatrow}
\usepackage{wrapfig}

\begin{document}
\title{PAN: Projective Adversarial Network for Medical Image Segmentation}
\titlerunning{PAN}
%
\author{Naji Khosravan\inst{2}, Aliasghar Mortazi\inst{2}, Michael Wallace\inst{1}, Ulas Bagci\inst{2}}
\authorrunning{N. Khosravan et al.}
%
\institute{Mayo Clinic Cancer Center, Jacksonville, FL.\and Center for Research in Computer Vision (CRCV),\\ School of Computer Science,
University of Central Florida, Orlando, FL.}
\maketitle              
\begin{abstract}
Adversarial learning has been proven to be effective for capturing long-range and high-level label consistencies in semantic segmentation. Unique to medical imaging, capturing 3D semantics in an effective yet computationally efficient way remains an open problem. In this study, we address this computational burden by proposing a novel projective adversarial network, called PAN, which incorporates high-level 3D information through 2D projections. Furthermore, we introduce an attention module into our framework that helps for a selective integration of global information directly from our \textit{segmentor} to our adversarial network. For the clinical application we chose pancreas segmentation from CT scans. Our proposed framework achieved state-of-the-art performance without adding to the complexity of the segmentor.

\end{abstract}

\section{Introduction} 
Segmentation has been a major area of interest within the fields of computer vision and medical imaging for years. Owing to their success, deep learning based algorithms have become the standard choice for semantic segmentation in the literature. Most state-of-the-art studies model segmentation as a pixel-level classification problem~\cite{deeplab_crf,rethinking_atrous,enc_dec_atrous}. Pixel-level loss is a promising direction but, it fails to incorporate global semantics and relations. To address this issue researchers have proposed a variety of strategies. A great deal of previous research uses a post-processing step to capture pairwise or higher level relations. Conditional Random Field (CRF) was used in~\cite{deeplab_crf} as an offline post-processing step to modify edges of objects and remove false positives in CNN output. In other studies, to avoid offline post-processing and provide an end-to-end framework for segmentation, mean-field approximate inference for CRF with Gaussian pairwise potentials was modeled through Recurrent Neural Network (RNN)~\cite{rnn_crf}. 

In parallel to post processing attempts, another branch of research tried to capture this global context through multi-scale or pyramid frameworks. In~\cite{deeplab_crf,rethinking_atrous,enc_dec_atrous}, several spatial pyramid pooling at different scales with both conventional convolution layers and \textit{Atrous} convolution layers were used to keep both contextual and pixel-level information. Despite such efforts, combining local and global information in an optimal manner is not a solved problem, yet.

Following by the seminal work by Goodfellow et.al. in~\cite{goodfellow2014generative} a great deal of research has been done on adversarial learning~\cite{gan_review,luc2016semantic, segan, conditionalGan}. Specific to segmentation, for the first time, Luc et. al.~\cite{luc2016semantic} proposed the use of a discriminator along with a segmentor in an adversarial min-max game to capture long-range label consistencies. In another study \textit{SegAN} was introduced, in which the segmentor plays the role of generator being in a min-max game with a discriminator with a multi-scale~\textit{L1} loss~\cite{segan}. A similar approach was taken for structure correction in chest X-rays segmentation in~\cite{scan}. A conditional GAN approach was taken in~\cite{conditionalGan} for brain tumor segmentation. 

In this paper, we focused on the challenging problem of pancreas segmentation from CT images, although our framework is generic and can be applied to any 3D object segmentgation problem. This particular application has unique challenges due to the complex shape and orientation of pancreas, having low contrast with neighbouring tissues and relatively small and varying size. Pancreas segmentation were studied widely in the literature. Yu et al. introduced a recurrence saliency transformation network, which uses the information from previous iteration as a spatial weight for current iteration~\cite{P_recurrent}. In another attempt, U-Net with an attention gate was proposed in~\cite{P_attention}. Similarly, a two-cascaded-stage based method was used to localize and segment pancreas from CT scans in~\cite{P_spatial}. A prediction-segmentation mask was used in~\cite{P_fixed} for constraining the segmentation with a coarse-to-fine strategy. Furthermore, a segmentation network with RNN was proposed in~\cite{P_rnn} to capture the spatial information among slices. The unique challenges of pancreas segmentation (complex shape and small organ) shifted the literature towards methods with coarse-to-fine and multi-stage frameworks, promising but computationally expensive.

\textbf{Summary of our contributions:} The current literature on segmentation fails to capture 3D high-level shape and semantics with a low-computation and effective framework. In this paper, for the fist time in the literature, we propose a projective adversarial network (PAN) for segmentation to fill this research gap. Our method is able to capture 3D relations through 2D projections of objects, without relying on 3D images or adding to the complexity of the segmentor. Furthermore, we introduce an attention module to selectively integrate high-level, whole-image features from the \textit{segmentor} into our adversarial network. With comprehensive evaluations, we showed that our proposed framework achieves the state-of-the-art performance on publicly available CT pancreas segmentation dataset~\cite{P_deeporgan} even when a simple encoder-decoder network was used as \textit{segmentor}.

\section{Method}
\begin{figure}[h]
\centering
\includegraphics[scale=0.6]{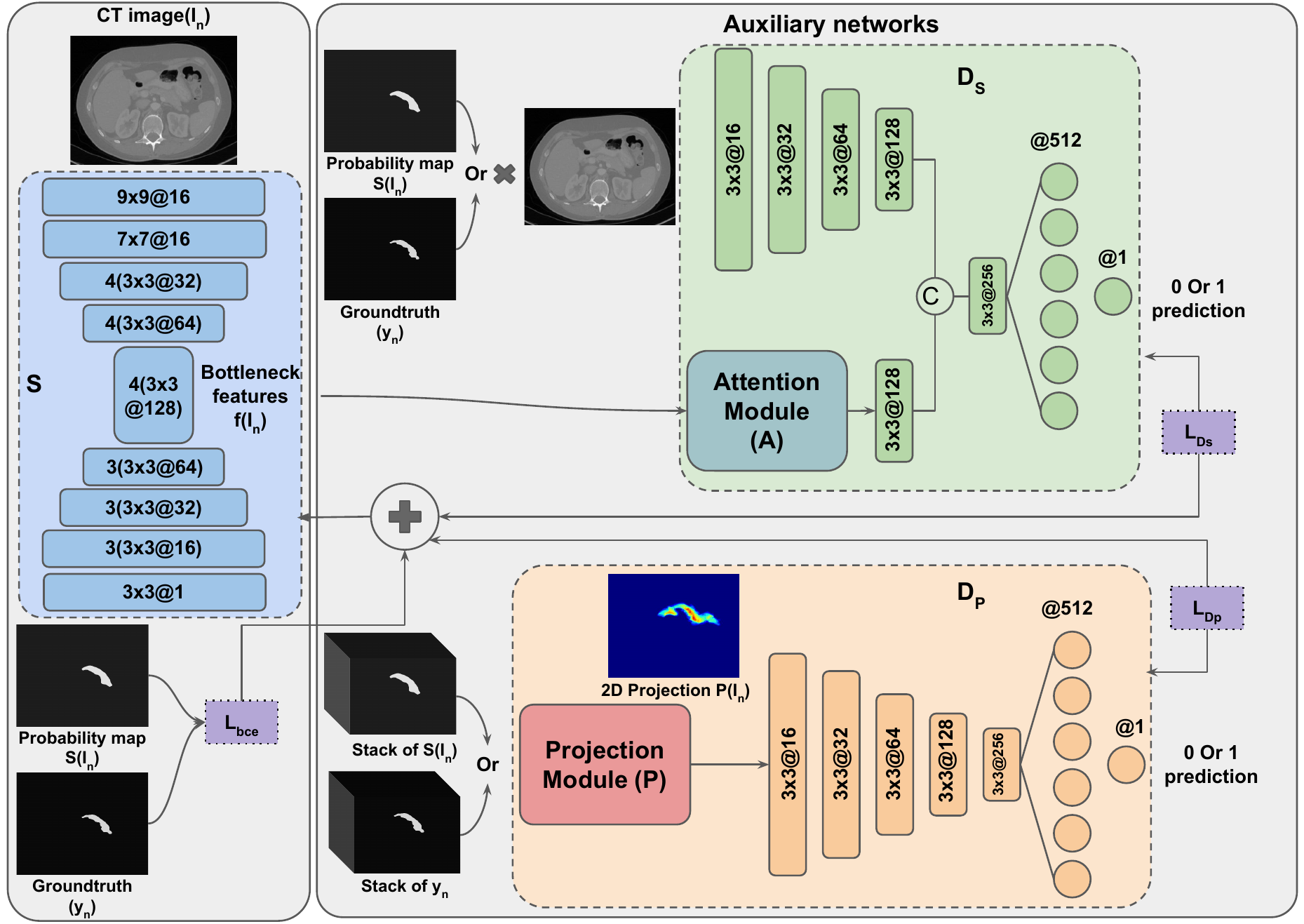}
\caption{The proposed framework consists of a segmentor $S$ and two adversarial networks, $D_{s}$ and $D_{p}$. $S$ was trained with a hybrid loss from $D_{s}$, $D_{p}$ and the ground-truth.\label{fig:sys}}
\end{figure}

Our proposed method is built upon the adversarial networks. The proposed framework's overview is illustrated in Figure~\ref{fig:sys}. We have three networks: a segmentor ($S$ in Figure~\ref{fig:sys}), which is our main network and was used during the test phase, and two adversarial networks ($D_{s}$ and $D_{p}$ in Figure~\ref{fig:sys}), each with a specific task. The first adversarial network ($D_{s}$) captures high-level \textit{spatial} label contiguity while the second adversarial network ($D_{p}$) enforces the \textit{3D semantics} through a 2D projection learning strategy. The adversarial networks were used only during the training phase to boost the performance of the segmentor without adding to its complexity.
\vspace{-.7cm}
\vspace{-.8cm}

\subsection{Segmentor (S)}
Our base network is a simple fully convolutoinal network with an encoder-decoder architecture. The input to the segmentor is a 2D grey-scale image and the output is a pixel-level probability map. The probability map shows probability of presence of the object at each pixel. 
We use a hybrid loss function (explained in details in Section~\ref{Adversarial_t}) to update the parameters our segmentor ($S$). This loss function is composed of three terms enforcing: (1) pixel-level high-resolution details, (2) spatial and high-range label continuity, (3) 3D shape and semantics, through our novel projective learning strategy. 

As can be seen in Figure~\ref{fig:sys}, the segmentor contains $10$ conv layers in the encoder, $10$ conv layers in the decoder and $4$ conv layers as the bottleneck. The last conv layer is a $1\times 1$ conv layer with the channel output of $1$, combining channel-wise information in the highest scale. This layer is followed by a sigmoid function to create the notion of probability. 

\subsection{Adversarial Networks}\label{att_module} 
Our adversarial networks are designed with the goal of compensating for the missing global relations and correcting higher-order inconsistencies, produced by a single pixel-level loss. Each of these networks produces an adversarial signal and apply it to the segmentor as a term in the overall loss function (Equation~\ref{hybrid}). The details of each network is described below:\\

\vspace{-.1cm}
\noindent\textbf{Spatial semantics network ($\textbf{D}_{\textbf{s}}$):} This network is designed to capture spatial consistencies within each frame. The input to this network is either the segmented object by the ground-truth or by the segmentor's prediction. The Spatial semantics network ($D_{s}$) is trained to discriminate between these two inputs with a binary cross-entropy loss, formulated as in Equation~\ref{spatial}. The adversarial signal produced by the negative loss of $D_{s}$ to $S$ forces $S$ to produce predictions closer to ground-truth in terms of spatial semantics.

As illustrated in Figure~\ref{fig:sys} top right, $D_{s}$ has a two-branch architecture with a late fusion. The top branch processes the segmented objects by ground-truth or segmentor's prediction. We propose an extra branch of processing, getting the bottleneck features corresponding to the original gray-scale input image, and passing them to an attention module for an information selection. The processed features are then concatenated with the first branch and passed through the shared layers. We believe that having the high-level features of whole image along with the segmentations improves the performance of $D_{s}$.

Our attention module learns where to attend in the feature space to have a more discriminative information selection and processing. The details of the attention module are described in the following.\\

\begin{wrapfigure}{r}{0.41\textwidth}
\vspace{-.9cm}
\centering
\includegraphics[scale=0.4]{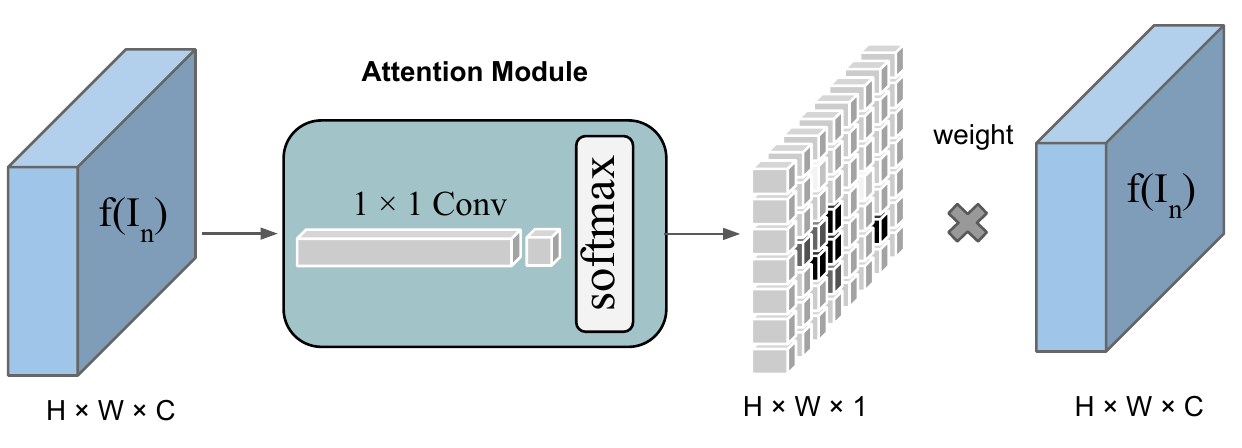}
\caption{Attention module assigns a weight to each feature allowing for a soft selection of information.\label{fig:att}}
\end{wrapfigure}
\noindent\textbf{Attention module ($\textbf{A}$):} We feed the high-level features form the segmentor's bottleneck to $D_{s}$. These features contain global information about the whole frame. We use a soft-attention mechanism, in which our attention module assigns a weight to each feature based on its importance for discrimination. The attention module gets the features with shape $w\times h\times c$, as input, and outputs a weight set with a shape of $w\times h\times 1$. $A$ is composed of two $1\times 1$ convolution layers followed by a softmax layer (Figure~\ref{fig:att}). The softmax layer introduces the notion of \textit{soft selection} to this module. The output of $A$ is then multiplied to the features before being passed to the rest of the network.\\

\noindent\textbf{Projective network ($\textbf{D}_{\textbf{p}}$):} Any 3D object can be projected into 2D planes from specific viewpoints, resulting in multiple 2D images. The 2D projection contains 3D semantics information, to be retrieved. In this section, we introduce our projective network ($D_{p}$). The main task of $D_{p}$ is to capture 3D semantics without relying on 3D data and from the 2D projections. Inducing 3D shapes form 2D images has previously been done for 3D shape generation~\cite{gadelha20173d}. Unlike existing notions, however, in this paper we propose 3D semantics induction from 2D projections, to benefit segmentation for the first time in the literature. 

The projection module ($P$) projects a 3D volume (V) on a 2D plane as:
\begin{equation} \label{projection}
    P((i,j),V) = 1 - \exp^{-\sum_{k}V(i,j,k)},
\end{equation}
where each pixel in the 2D projection $P((i,j),V)$ gets a value in the range of $[0,1]$ based on the number of voxel occupancy in the third dimension of corresponding $3D$ volume ($V$). For the sake of simplicity, we refer to the projection of a 3D volume $V$ as $P(V)$.
We pass each 3D image through our segmentor ($S$) slice by slice and stack the corresponding prediction maps. Then, these maps are fed to the projection module ($P$) and are projected in the axial view. 

The input to $D_{p}$ is either the projected ground-truth or projected prediction map produced by $S$. $D_{p}$ is trained to discriminate these inputs using the loss function defined in Equation~\ref{projective}. The adversarial term produced by $D_{p}$ in Equation~\ref{hybrid} forces $S$ to create predictions which are closer to ground-truth in terms of 3D semantics. Incorporating $D_{p}$ as an adversarial network to our segmentation framework helps $S$ to capture 3D information through a very simple 2D architecture and without adding to its complexity in the test time.

\subsection{Adversarial training} \label{Adversarial_t}
To train our framework, we use a hybrid loss function, which is a weighted sum of three terms. For a dataset of $N$ training samples of images and ground truths $(I_{n},y_{n})$, we define our hybrid loss function as:

\begin{equation} \label{hybrid}
    l_{hybrid} = \sum_{n=1}^{N} l_{bce}(S(I_{n}),y_{n}) - \lambda l_{D_{s}} - \beta l_{D_{p}},
\end{equation}
where $l_{D_{s}}$ and $l_{D_{p}}$ are the losses corresponding to $D_{s}$ and $D_{p}$ and $S(I_{n})$ is the segmentor's prediction. The first term in Equation~\ref{hybrid} is a weighted binary cross-entropy loss. This loss is the state-of-the-art loss function for semantic segmentation and for a grey-scale image $I$ with size $H\times W\times 1$ is defined as:

\begin{equation} \label{bce}
    l_{bce}(\Hat{y},y) = - \sum_{i=1}^{H\times W} (wy_{i}\log{\Hat{y}_{i}} + (1 - y_{i})\log{(1 - \Hat{y}_{i})}),
\end{equation}
where $w$ is the weight for positive samples, $y$ is the ground-truth label and $\Hat{y}$ is the network's prediction. Equation~\ref{bce} encourages $S$ to produce predictions similar to ground-truth and penalizes each pixel \textit{independently}. High-order relations and semantics cannot be captured by this term.

To account for this drawback, the second and third terms are added to train our auxiliary networks. $l_{D_{s}}$ and $l_{D_{p}}$ are defined below, respectively:

\begin{eqnarray} \label{spatial}
    l_{D_{s}} = l_{bce}(D_{s}(I_{n},y_{n}),1) + l_{bce}(D_{s}(I_{n},S(I_{n})),0),\\ \label{projective}
        l_{D_{p}} = l_{bce}(D_{p}(P_{I_{n}},P{y_{n}}),1) + l_{bce}(D_{p}(P_{I_{n}},P_{S(I_{n})}),0).
\end{eqnarray}


Here $P$ is the projection module, $l_{bce}$ is the binary cross-entropy loss with $w=1$ in Equation \ref{bce} corresponding to a single number ($0$ or $1$) as the output.


\section{Experiments and Results}
We evaluated the efficacy of our proposed system with the challenging problem of pancreas segmentation. This particular problem was selected due to the complex and varying shape of pancreas and relatively more difficult nature of the segmentation problem compared to other abdominal organs. In our experiments we show that our proposed framework outperforms other state-of-the-art methods and captures the complex 3D semantics with a simple encoder-decoder. Furthermore, we have created an extensive comparison to some baselines, designed specifically to show the effects of each block of our framework.

\noindent\underline{\textbf{Data and evaluation:}} We used the publicly available TCIA CT dataset from NIH~\cite{P_deeporgan}. This dataset contains a total of $82$ CT scans. The resolution of scans is $512\times 512\times Z$, $Z \in [181,466]$ is the number of slices in the axial plane. The voxels spacing ranges from $0.5mm$ to $1.0mm$. We used a randomly selected set of $62$ images for training and $20$ for testing to perform a 4-fold cross-validation. Dice Similarity Coefficient (DSC) is used as the metric of evaluation.
 
\noindent\underline{\textbf{Comparison to baselines:}}
\begin{wraptable}{r}{5.0cm}
\vspace{-1cm}
\caption{Comparison with baselines.}
\begin{tabular}{|l|l|c|}
\hline
\rowcolor{lightgray} 
\cellcolor{lightgray} & \textbf{Model} & \textbf{DSC\%} \\ \cline{2-3} 
\cellcolor{lightgray} &  Encoder-decoder (S)    &  57.7   \\ \cline{2-3} 
\cellcolor{lightgray} &  Atrous pyramid     &   48.2  \\ \cline{2-3} 
\cellcolor{lightgray}  &   $S + D_{s}$   &   85.0  \\ \cline{2-3} 
\cellcolor{lightgray}  &   $S + D_{s} + A$  &   85.9  \\ \cline{2-3} 
\multirow{-5}{*}{\cellcolor{lightgray}\rotatebox[origin=c]{90}{\tiny\textbf{1-fold}}} &     $S + D_{s} + A + D_{p}$ & \textbf{86.8}    \\ \hline
\end{tabular}
\label{table:baselines}
\end{wraptable}
To show the effect of each building block of our framework we designed an extensive set of experiments. In our experiments we start from only training a single segmentor (S) and go to our final proposed framework. Furthermore, we show comparison of encoder-decoder architecture with other state-of-the-art semantic segmentation architectures.

Table~\ref{table:baselines} shows the results adding of each building block of our framework. The eccoder-decoder architecture is the one showed in Figure~\ref{fig:sys} as $S$, while the Atrous pyramid architecture is similar to the recent work of~\cite{enc_dec_atrous}. This architecture is currently state-of-the-art for semantic segmentation. In which an Atrous pyramid is used to capture global context. We added an Atrous pyramid with $5$ different scales: $4$ Atrous convolutions at rates of $1,2,6,12$, with the global image pooling. We also replaced the decoder with $2$ simple upsampling and conv layers similar to the main paper~\cite{enc_dec_atrous}. We refer the readers to the main paper for more details about this architecture due to space limitations~\cite{enc_dec_atrous}.
We found out having an extensive processing in the decoder improves the results compared to the Atrous pyramid architecture (possibly a better choice for segmentation of objects at multiple scales). This is because our object of interest is relatively small. 

Moreover, we showed that adding a spatial adversarial notwork ($D_{s}$) can boost the performance of $S$ dramatically, in our task. Introducing attention ($A$) helps for a better information selection (as described in section~\ref{att_module}) and boosts the performance further. Finally, our best results is achieved by adding the projective adversarial network ($D_{p}$), which adds integration of 3D semantics into the framework. This supports our hypothesis that our segmentor has enough capacity in terms of parameters to capture all this information and with proper and explicit supervision can achieve state-of-the-art results. 

\noindent\underline{\textbf{Comparison to the state-of-the-art:}}
We provide the comparison of our method's performance with current state-of-the-art literature on the same TCIA CT dataset for pancreas segmentation. As can be seen from experimental validation, our method outperforms the state-of-the-art with dice scores, provides better efficiency (less computational burden). Of a note, the proposed algorithm's least achievement is consistently higher than the state of the art methods.

\vspace{-.6cm}
\begin{table}[h]
\caption{Comparison with state-of-the-art on TCIA dataset.}
\begin{tabular}{|l|l|c|c|c|}
\hline
\rowcolor{lightgray} 
\cellcolor{lightgray} & \textbf{Approach} & \textbf{Average DSC\%} & \textbf{Max DSC\%} & \textbf{Min DSC\%} \\ \cline{2-5} 
\cellcolor{lightgray} &  Roth et al.\cite{P_deeporgan}&  $71.42\pm 10.11$  & 86.29 & 23.99  \\ \cline{2-5}
\cellcolor{lightgray} &  Roth et al.\cite{roth2016spatial}&  $78.01\pm 8.20$  & 88.65 & 34.11  \\ \cline{2-5} 
\cellcolor{lightgray} &  Roth et al.\cite{P_spatial}&  $81.27\pm 6.27$  & 88.96 & 50.69  \\ \cline{2-5} 
\cellcolor{lightgray} &  Zhou et al.\cite{P_fixed} &  $82.37\pm 5.68$  & 90.85 & 62.43  \\ \cline{2-5} 
\cellcolor{lightgray} &  Cai et al.\cite{P_rnn} &  $82.40\pm 6.70$   & 90.10  & 60.00   \\ \cline{2-5} 
\cellcolor{lightgray} &  Yu et al.\cite{P_recurrent}&  $84.50\pm 4.97$  & \textbf{91.02} & 62.81  \\ \cline{2-5} 
\multirow{-8}{*}{\cellcolor{lightgray}\rotatebox[origin=c]{90}{\tiny\textbf{4-fold cross validation}}} & \textbf{Ours} & $\textbf{85.53}\pm \textbf{1.23}$ & 88.71 & \textbf{83.20}  \\ \hline
\end{tabular}
\label{table:soa}
\end{table}
\vspace{-1cm}

\section{Conclusion}
In this paper we proposed a novel adversarial framework for 3D object segmentation. We introduced a novel projective adversarial network, inferring 3D shape and semantics form 2D projections. The motivation behind our idea is that integration of 3D information through a fully 3D network, having all slices as input, is computationally infeasible. Possible workarounds are: 1)down-sampling the data or 2)sacrificing number of parameters, which are sacrificing information or computational capacity, respectively. We also introduced an attention module to selectively pass whole-frame high-level feature from the segmentor's bottleneck to the adversarial network, for a better information processing. We showed that with proper and guided supervision through adversarial signals a simple encoder-decoder architecture, with enough parameters, achieves state-of-the-art performance on the challenging problem of pancreas segmentation.
We achieved a dice score of $\textbf{85.53\%}$, which is state-of-the art performance on pancreas segmentation task, outperforming previous methods. Furthermore, we argue that our framework is general and can be applied to any 3D object segmentation problem and is not specific to a single application.

\bibliographystyle{splncs03}
\bibliography{sssd}
\end{document}